\pdfoutput=1
\documentclass[11pt]{article}

\usepackage[final]{acl}

\usepackage{tabularx}
\usepackage{times}
\usepackage{latexsym}
\usepackage[T1]{fontenc}
\usepackage[utf8]{inputenc}
\usepackage{amsmath}
\usepackage{booktabs}
\usepackage{multirow}
\usepackage{microtype}
\usepackage{inconsolata}
\usepackage{graphicx}
\usepackage{pgfplots}
\pgfplotsset{compat=1.17}
\usepackage{enumitem}
\usepackage{color}
\usepackage{soul}
\usepackage{array}
\usepackage{placeins}
\usepackage{url}
\usepackage{hyperref}

\DeclareUnicodeCharacter{2212}{-}

\title{When Does Continual Pre-Training Help? \\
A Controlled Study of Dialectal Adaptation for Québec French}

\author{
  Eeham Khan$^1$ \and Richard Khoury$^2$ \and Leila Kosseim$^1$ \\
  $^1$CLaC Lab, Department of Computer Science and Software Engineering, Concordia University \\
  $^2$GRAAL, Department of Computer Science and Software Engineering, Universit\'e Laval \\
  \texttt{\{eeham.khan, leila.kosseim\}@concordia.ca, richard.khoury@ift.ulaval.ca}
}

\begin{document}
\maketitle

\begin{abstract}
Continual pre-training (CPT) is widely used to adapt pretrained language models to dialectal and non-standard language varieties, yet it remains unclear when this additional training is necessary, what drives its benefits, and whether aggregate evaluation scores adequately capture its effects. In this paper, we present a controlled study of CPT for Québec French language understanding across nine LLMs spanning different families and scales. Using a budget-matched protocol of 30M tokens per experiment, our results show that the benefit of CPT depends strongly on the starting model’s existing competence in the target variety. When CPT is beneficial, the improvements are driven more by corpus composition than by data volume, with literary text accounting for most of the gain. Paired Québec and standard French evaluations further show that CPT primarily improves expression comprehension rather than grammatical competence, and that corpus register determines which regional expressions are acquired. These gains also involve capability-specific trade-offs, including declines in reading comprehension that aggregate scores can conceal. Overall, these findings show that CPT for dialectal adaptation should not be treated as a uniformly beneficial default. Instead, adaptation strategies should begin with a careful assessment of the base model's existing competence, accompanied by source-level data analysis and task-level evaluation.
\end{abstract}

\section{Introduction}
\label{sec:intro}

Large Language Models (LLMs) are pretrained on massive corpora to support a wide range of applications, including summarization, natural language and code generation, and open-domain dialogue. Despite their capabilities, LLMs often perform better on the standard form of a language than on its regional varieties. This phenomenon, known as the \emph{dialect gap} \citep{kantharuban-etal-2023-quantifying}, persists even in multilingual models trained on large and diverse corpora \citep{openai2026openaio1card,Comanici2025Gemini2P,lin2024mala500}: when a dialect is underrepresented in the training data, its linguistic capabilities are often reduced in the resulting LLM.

Continual pre-training (CPT) \citep{ke2023continual} is widely used to address dialectal adaptation. By taking an existing model pretrained on the standard language and continuing its training on dialectal text, CPT aims to improve regional language competency while preserving the broad capabilities learned during pre-training \citep{gururangan-etal-2020-dont,sarkar2022parameter}. 

However, this approach leaves several questions unresolved: 1) Base models can differ substantially in their existing competence in the target variety, making adaptation unnecessary for models that already perform well; 2) dialectal corpora are often small and heterogeneous, combining a variety of genres and registers. Unless the amount of training data is controlled, it is difficult to determine whether performance changes are caused by specific genres and registers or by greater volume of texts; and 3) aggregate benchmark scores may not be sufficiently representative of the gains (or degradation) of an LLM's capabilities in specific tasks. We isolate three factors that are usually confounded in dialectal adaptation studies: the base model’s existing competence, the composition of the adaptation corpus, and the granularity of evaluation. We focus on Québec French, for which paired dialectal and standard French benchmarks targeting the same capabilities are available, which makes this isolation possible. Our evaluation is restricted to language understanding; whether CPT changes the dialectal character of generated text is left for future work.

To investigate these research questions, we conduct a controlled, budget-matched study of CPT for Québec French. Our evaluation spans nine LLMs from three model families (0.5B to 14B parameters), with each experiment utilizing exactly 30M training tokens. This uniform constraint not only enables direct comparisons between various corpus compositions, but also mirrors the realistic data scarcity inherent to low-resource dialectal adaptation, where clean, target-variety text is often difficult to obtain. We compare every adapted model with its own unadapted baseline. Evaluation pairs each Québec French benchmark with a standard French benchmark measuring the same linguistic capability, drawn from the COLE suite \citep{cole}, alongside standard French retention tasks and held-out perplexity scores. 

Together, our findings suggest that, in the low-resource regime where dialectal adaptation is typically performed, CPT should not be treated as the default response to dialect under-representation. Under a limited adaptation budget, the impact of CPT depends on the starting model, the composition of the adaptation corpus, and trade-offs among individual capabilities that aggregate scores and perplexity may fail to reveal. Specifically, our contributions in this work are:
\begin{itemize}[leftmargin=*, itemsep=0.5pt, topsep=1pt]
\item We show that the benefit of dialectal CPT for comprehension depends strongly on the base model’s existing competence in the target variety, and that informal-only adaptation degrades regional expression comprehension in eight of nine models.

\item We show that corpus composition matters more than data volume, with gains driven primarily by literary text on CroissantLLM, the model that gains most from CPT.

\item We show that aggregate evaluation scores and held-out perplexity provide an incomplete evaluation of CPT: task-level evaluation reveals large and opposing capability changes, including losses in reading comprehension, that remain hidden behind aggregate performance.
\end{itemize}

\section{Related Work}
\label{sec:related}

\subsection{Continual Pre-Training (CPT)}

Continual pre-training (CPT) adapts pretrained language models to new domains via self-supervised learning on unlabeled text \citep{gururangan-etal-2020-dont,ke2023continual}. Recent work explores strategies such as multilingual training \citep{csaki-etal-2024-sambalingo,fujii2024continual,joshi-etal-2025-adapting} and data replay \citep{gupta2023continual,ibrahim2024simple} to improve adaptation while preserving prior knowledge. Crucially, CPT's efficacy depends heavily on model scale and domain size: small or mismatched adaptation corpora can leave stronger models unchanged or even degrade them \citep{reid-etal-2022-m2d2}.

\subsection{Dialectal and Low-Resource Adaptation}

Because LLMs perform unevenly on under-represented regional varieties \citep{kantharuban-etal-2023-quantifying,dialectbench}, a standard approach is to adapt existing models rather than train them from scratch. While low-resource adaptation often applies CPT to massive, heterogeneous multilingual corpora \citep{lin2024mala500,emma500,afriquellm}, dialectal adaptation is distinct because models already possess substantial knowledge of the associated standard language. Prior work employs techniques such as supervised fine-tuning, LoRA, and synthetic data to bridge these dialect gaps \citep{liu-etal-2023-dada,bafna-etal-2025-dialup,lora}. For Québec French specifically, recent work has demonstrated that adaptation is feasible under strict resource constraints, while highlighting that outcomes remain highly sensitive to corpus composition \citep{khan-etal-2026-low}. 

To our knowledge, no previous work has systematically isolated the role of base model competence, demonstrated how specific textual genres drive the acquisition of regional expressions rather than grammar, or studied how standard aggregate metrics mask the task-specific capability trade-offs inherent to dialectal continual pre-training.

\section{Corpus and Adaptation}
\label{sec:setup}
\subsection{Québec French Corpus}
\label{sec:corpus}
To investigate these research questions, we conduct a controlled, budget-matched study of CPT for Québec French. We use the mixed-source Québec French corpus of \cite{khan-etal-2026-low} which covers multiple genres and registers. Table~\ref{tab:corpus} shows the composition of the corpus. 

\begin{table}[t]
\centering
\caption{Composition of the Mixed-Source Québec French Adaptation Corpus. Token count was computed using the CroissantLLM tokenizer\protect\footnotemark[1].}
\vspace{-2.5mm}
\label{tab:corpus}
\small
\setlength{\tabcolsep}{4pt}
\begin{tabular}{l l l r}
\toprule
\textbf{Source} & \textbf{Genre} & \textbf{Register} & \textbf{Tokens} \\
\midrule
BEQ & Literary & Formal & 15.4M \\
Québec Wikipedia & Encyclopedic & Formal & 31.2M \\
Le Soleil & News & Formal & 5.9M \\
\cmidrule(lr){1-4}
\multicolumn{3}{l}{\emph{Formal subtotal}} & 52.5M \\
\midrule
CRIFUQ & Speech & Informal & 0.8M \\
Facebook & Social media & Informal & 2.4M \\
Dépotoir & Web forum & Informal & 22.8M \\
MontrealRacing & Web forum & Informal & 4.2M \\
YouTube & Comments & Informal & 2.2M \\
Reddit & Social media & Informal & 1.8M \\
\cmidrule(lr){1-4}
\multicolumn{3}{l}{\emph{Informal subtotal}} & 34.2M \\
\midrule
\multicolumn{3}{l}{\textbf{Total}} & 86.7M \\
\bottomrule
\end{tabular}
\end{table}
\footnotetext[1]{available at \url{https://huggingface.co/croissantllm}.}

As shown in Table~\ref{tab:corpus}, the complete 86.7M-token corpus is drawn from nine sources categorized into two primary registers. The \emph{formal} register (52.5M tokens) consists of clean, standardized written language, including public-domain literature (\textit{Bibliothèque Électronique du Québec}), Québec Wikipedia and news articles. Conversely, the \emph{informal} register (34.2M tokens) captures vernacular language, comprising transcribed speech (\textit{Centre de recherche interuniversitaire sur le français en usage au Québec}) alongside colloquial user-generated web text from regional discussion forums and social media platforms. % leila removed this: Together, these two registers provide a distinct contrast in textual genre.

\subsection{Adaptation Setup}

\begin{figure*}[t]
\centering
\scalebox{.9}{ % added by leila
\begin{tikzpicture}

% --- ROW 1: GRAMMAR (Syntax) ---

\begin{axis}[
  name=qfrblimpax,
  width=0.49\textwidth,
  height=4.5cm, 
  ybar,
  bar width=3.4pt,
  title={\footnotesize (a) Qu\'ebec French grammar: $\Delta$ QFrBLiMP},
  ylabel={$\Delta$ Macro-F1 vs.\ base},
  symbolic x coords={Croissant 1.35B,Qwen 0.5B,Qwen 1.5B,Llama 1B,Qwen 3B,Llama 3B,Qwen 7B,Llama 8B,Qwen 14B},
  xtick=data,
  x tick label style={rotate=45, anchor=east, font=\scriptsize},
  ymin=-9, ymax=11,
  ymajorgrids,
  grid style={dashed,gray!25},
  tick label style={font=\scriptsize},
  label style={font=\scriptsize},
  legend style={font=\scriptsize, at={(0.98,0.98)}, anchor=north east, draw=none},
  legend cell align=left,
]
\addplot+[fill=blue!60, draw=blue!60!black] coordinates {(Croissant 1.35B,0.7) (Qwen 0.5B,-0.7) (Qwen 1.5B,1.4) (Llama 1B,0.8) (Qwen 3B,-0.4) (Llama 3B,-0.3) (Qwen 7B,0.4) (Llama 8B,-1.4) (Qwen 14B,0.1)};
\addlegendentry{Formal-only CPT}
\addplot+[fill=red!55, draw=red!60!black] coordinates {(Croissant 1.35B,-0.8) (Qwen 0.5B,-2.0) (Qwen 1.5B,0.4) (Llama 1B,-0.4) (Qwen 3B,-1.3) (Llama 3B,-0.7) (Qwen 7B,0.0) (Llama 8B,0.1) (Qwen 14B,-0.7)};
\addlegendentry{Informal-only CPT}
\addplot[black, thick, forget plot] coordinates {(Croissant 1.35B,0) (Qwen 14B,0)};
\end{axis}

\begin{axis}[
  name=mfrblimpax,
  at={(qfrblimpax.outer east)}, anchor=outer west, xshift=6pt,
  width=0.49\textwidth,
  height=4.5cm, 
  ybar,
  bar width=3.4pt,
  title={\footnotesize (b) Standard French grammar: $\Delta$ MultiBLiMP-Fr},
  symbolic x coords={Croissant 1.35B,Qwen 0.5B,Qwen 1.5B,Llama 1B,Qwen 3B,Llama 3B,Qwen 7B,Llama 8B,Qwen 14B},
  xtick=data,
  x tick label style={rotate=45, anchor=east, font=\scriptsize},
  ymin=-9, ymax=11,
  ymajorgrids,
  grid style={dashed,gray!25},
  tick label style={font=\scriptsize},
  label style={font=\scriptsize},
  legend style={font=\scriptsize, at={(0.98,0.98)}, anchor=north east, draw=none},
  legend cell align=left,
]
\addplot+[fill=blue!60, draw=blue!60!black] coordinates {(Croissant 1.35B,-1.3) (Qwen 0.5B,-4.8) (Qwen 1.5B,1.7) (Llama 1B,-0.9) (Qwen 3B,0.0) (Llama 3B,-1.3) (Qwen 7B,0.0) (Llama 8B,0.0) (Qwen 14B,-1.7)};
\addlegendentry{Formal-only CPT}
\addplot+[fill=red!55, draw=red!60!black] coordinates {(Croissant 1.35B,-1.3) (Qwen 0.5B,1.7) (Qwen 1.5B,1.3) (Llama 1B,-2.2) (Qwen 3B,0.9) (Llama 3B,-2.2) (Qwen 7B,0.0) (Llama 8B,-0.1) (Qwen 14B,-3.2)};
\addlegendentry{Informal-only CPT}
\addplot[black, thick, forget plot] coordinates {(Croissant 1.35B,0) (Qwen 14B,0)};
\end{axis}

% --- ROW 2: LEXICON (Expressions) ---

\begin{axis}[
  name=coreax,
  at={(qfrblimpax.outer south west)}, anchor=outer north west, yshift=-0.2cm,
  width=0.49\textwidth,
  height=4.5cm, 
  ybar,
  bar width=3.4pt,
  title={\footnotesize (c) Qu\'ebec French expressions: $\Delta$ QFrCoRE},
  ylabel={$\Delta$ Macro-F1 vs.\ base},
  symbolic x coords={Croissant 1.35B,Qwen 0.5B,Qwen 1.5B,Llama 1B,Qwen 3B,Llama 3B,Qwen 7B,Llama 8B,Qwen 14B},
  xtick=data,
  x tick label style={rotate=45, anchor=east, font=\scriptsize},
  ymin=-9, ymax=11,
  ymajorgrids,
  grid style={dashed,gray!25},
  tick label style={font=\scriptsize},
  label style={font=\scriptsize},
]
\addplot+[fill=blue!60, draw=blue!60!black] coordinates {(Croissant 1.35B,8.2) (Qwen 0.5B,1.5) (Qwen 1.5B,0.6) (Llama 1B,1.4) (Qwen 3B,1.4) (Llama 3B,0.1) (Qwen 7B,1.5) (Llama 8B,-1.4) (Qwen 14B,-2.2)};
\addplot+[fill=red!55, draw=red!60!black] coordinates {(Croissant 1.35B,0.7) (Qwen 0.5B,-3.4) (Qwen 1.5B,-4.0) (Llama 1B,-4.3) (Qwen 3B,-4.3) (Llama 3B,-6.1) (Qwen 7B,-3.8) (Llama 8B,-3.3) (Qwen 14B,-7.8)};
\addplot[black, thick, forget plot] coordinates {(Croissant 1.35B,0) (Qwen 14B,0)};
\end{axis}

\begin{axis}[
  name=mfrcoeax,
  at={(coreax.outer east)}, anchor=outer west, xshift=6pt,
  width=0.49\textwidth,
  height=4.5cm, 
  ybar,
  bar width=3.4pt,
  title={\footnotesize (d) Standard French expressions: $\Delta$ MFrCoE},
  symbolic x coords={Croissant 1.35B,Qwen 0.5B,Qwen 1.5B,Llama 1B,Qwen 3B,Llama 3B,Qwen 7B,Llama 8B,Qwen 14B},
  xtick=data,
  x tick label style={rotate=45, anchor=east, font=\scriptsize},
  ymin=-9, ymax=11,
  ymajorgrids,
  grid style={dashed,gray!25},
  tick label style={font=\scriptsize},
  label style={font=\scriptsize},
]
\addplot+[fill=blue!60, draw=blue!60!black] coordinates {(Croissant 1.35B,2.0) (Qwen 0.5B,0.7) (Qwen 1.5B,0.3) (Llama 1B,0.4) (Qwen 3B,-1.1) (Llama 3B,0.1) (Qwen 7B,-0.2) (Llama 8B,-0.3) (Qwen 14B,-0.9)};
\addplot+[fill=red!55, draw=red!60!black] coordinates {(Croissant 1.35B,4.8) (Qwen 0.5B,4.2) (Qwen 1.5B,5.3) (Llama 1B,2.2) (Qwen 3B,6.6) (Llama 3B,3.8) (Qwen 7B,6.1) (Llama 8B,5.7) (Qwen 14B,6.5)};
\addplot[black, thick, forget plot] coordinates {(Croissant 1.35B,0) (Qwen 14B,0)};
\end{axis}

\end{tikzpicture}
}
\caption{Register Sweep Across All Models: Shifts in grammatical performance (top row) and expression comprehension (bottom row) at the two register extremes for all nine models. Models are ordered left to right by baseline QFrCoRE. Results are difference in Macro-F1 compared to baseline result, averaged over three seeds.}
\vspace{-2.5mm}
\label{fig:extremes}
\end{figure*}
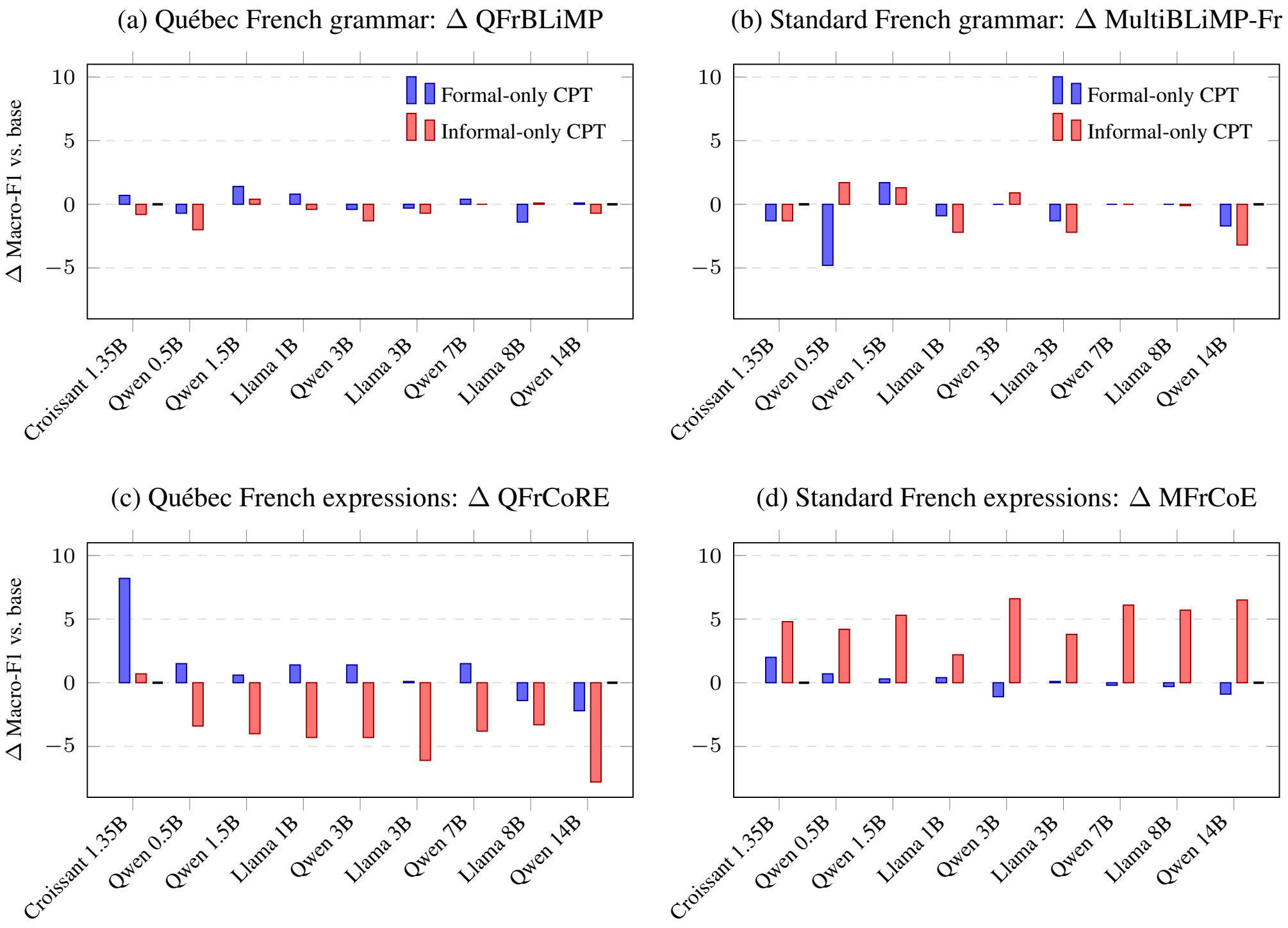

\paragraph{Models and Training.}
We perform CPT on nine LLMs across three families to compare adaptation across different architectures and levels of prior French exposure: CroissantLLM (1.35B)~\citep{faysse2024croissantllm}, Llama-3.2/3.1 (1B, 3B, 8B)~\citep{grattafiori2024llama3}, and Qwen2.5 (0.5B, 1.5B, 3B, 7B, 14B)~\citep{qwen25}. To ensure a fair comparison and avoid oversampling our smallest informal sources, all experiments are strictly budget-matched to 30M tokens. We adapt all models using LoRA while keeping the base-model weights frozen. Full optimization details, hyperparameters, and checkpoint selection criteria are provided in Appendix~\ref{sec:hyperparameters}.

\paragraph{Evaluation.}
We evaluate our models on two separate axes: dialect-specific acquisition and standard French retention. To isolate Quebec dialectal competence, we use two benchmarks: 
\begin{description} [noitemsep] % leila
    \item[QFrBLiMP \citep{beauchemin-etal-2026-qfrblimp}]~is~a grammatical acceptability task consisting of 1,761 sentence pairs, where the model must distinguish a correct sentence from one containing a common grammatical mistake sourced from official Québec guidelines.
    \item[QFrCoRE \citep{beauchemin-etal-2026-idiom}]~is~a definition-matching task containing 4,633 regional idiomatic multi-word expressions, requiring the model to identify the correct meaning from ten possible candidates. 
\end{description}

We pair these with standard French equivalents that target the same linguistic capabilities in order to serve as the standard French control:
\begin{description} [noitemsep] % leila
    \item[MultiBLiMP-Fr \citep{jumelet-etal-2026-multiblimp}]~is~a grammatical acceptability task parallel to QFrBLiMP, consisting of 255 sentence pairs.
    \item[MFrCoE \citep{beauchemin-etal-2026-idiom}]~is~a definition-matching task parallel to QFrCoRE, containing 4,938 standard French expressions.
\end{description}

In addition, to monitor for standard French forgetting during adaptation, we track performance on four general standard French tasks from the COLE suite \citep{cole}:
\begin{description}[noitemsep] % leila
    \item[AlloCiné \citep{blard2020allocine}] is a binary sentiment classification task evaluating a corpus of 20,000 French movie reviews.
    \item[PAWS-X \citep{yang2019pawsx}] is a paraphrase detection task requiring the model to identify whether two lexically overlapping sentences are semantically equivalent. We used the test set of 2,000 French samples for evaluation.
    \item[Fr-BoolQ \citep{clark2019boolq}] is a binary reading comprehension task consisting of yes/no questions paired with context passages. We used the French subset of 178 test samples.
    \item[MMS \citep{brand24mmsbenchmark}] is a sentiment analysis task evaluating text drawn from diverse domains using a binary or trinary classification schema. We used the French subset of 63,190 test samples.
\end{description}

\subsection{Experimental Setup}
\label{sec:results}

% \begin{table*}[t]
% \centering
% \caption{Continuous Register Sweep on CroissantLLM 1.35B: Results are Macro-F1, averaged over three seeds. Perplexity (PPL) is evaluated on held-out Québec French and standard French sets.}
% \vspace{-2.5mm}
% \label{tab:register}
% \footnotesize
% \setlength{\tabcolsep}{4pt}
% \begin{tabular}{@{} l cc cccc cc @{}}
% \toprule
% & \multicolumn{2}{c}{\textbf{Québec French}} 
% & \multicolumn{4}{c}{\textbf{Standard French}} 
% & \multicolumn{2}{c}{\textbf{PPL} $\downarrow$} \\
% \cmidrule(lr){2-3} \cmidrule(lr){4-7} \cmidrule(lr){8-9}
% \textbf{Condition} & QFrBLiMP & QFrCoRE & Allo. & PAWS & Fr-BoolQ & MMS & QC & FR \\
% \midrule
% \emph{Base (no CPT)} & 89.2 & 27.5 & 50.4 & 35.5 & 56.2 & 25.4 & 17.24 & 8.31 \\
% \midrule
% Formal-only   & 89.9 & 35.7 & 56.6 & 41.0 & 37.7 & 25.5 & 15.63 & 9.89 \\
% 75/25         & 89.4 & 33.9 & 54.1 & 39.1 & 40.2 & 25.4 & 15.79 & 9.73 \\
% Mixed (50/50) & 90.2 & 33.7 & 50.8 & 39.0 & 40.8 & 25.3 & 15.99 & 9.62 \\
% 25/75         & 91.0 & 32.2 & 57.2 & 38.6 & 41.7 & 25.3 & 16.30 & 9.53 \\
% Informal-only & 88.5 & 28.2 & 60.0 & 40.7 & 38.9 & 25.1 & 23.19 & 10.70 \\
% \midrule
% Double (60M)$^\dag$ & 90.7 & 33.8 & 61.8 & 39.0 & 43.5 & 25.4 & 15.67 & 9.84 \\
% \bottomrule
% \multicolumn{9}{@{}l}{\scriptsize $^\dag$60M tokens instead of 30M.}
% \end{tabular}
% \end{table*}

To investigate CPT for dialectal adaptation, we designed several experiments along three core dimensions: 1) the impact of corpus composition (see Section \ref{sec:results-rq1}); 2) the impact of base model capacity (see Section \ref{sec:results-rq2}); and 3) the masking effect of aggregate metrics (see Section \ref{sec:results-rq3}). 

Across all experiments, we report macro-F1 scored via length-normalized candidate log-likelihoods, averaged over three random seeds. Additionally, we compute held-out perplexity (PPL) for both language varieties. Unless otherwise noted, all performance changes are measured against each model's unadapted baseline.

\section{Impact of Corpus Composition}
\label{sec:results-rq1}

To evaluate how register and genre influence dialectal acquisition, we run three experiments. The first (see Section \ref{sec:register}) examines the formal-only and informal-only register extremes across all nine LLMs. The second experiment (see Section \ref{sec:croissant-sweep}) varies the formal-to-informal ratio exclusively on CroissantLLM to measure how granular changes in register impact the acquired dialect. Finally, the third experiment (see Section \ref{sec:composition-ablation}) conducts a leave-one-genre-out ablation relative to the full mixed-source corpus to isolate the contribution of specific genres.

Table~\ref{tab:corpus-matrix} shows the full mixed-source corpus and the variants obtained by reweighting individual sources. To vary the register, we construct five variants ranging from formal-only (100/0) to informal-only (0/100). Within each register, sources are sampled proportionally to their distribution in the full corpus. If a source's target allocation exceeds its raw availability (e.g., CRIFUQ), we do not artificially cycle the text over multiple epochs. Instead, all available tokens from that source are used exactly once, and the shortfall is backfilled with random sentences sampled from other sources within the same register. This protocol maintains the strict 30M-token budget across all experiments while preventing overfitting to the smallest domains.

\begin{table*}[t]
\centering
\caption{Corpus Variants Used in the Register Sweep. Values represent the allocated token budget (in millions) for each source. Sources within each register are sampled proportionally to their raw availability to maintain exactly 30M tokens per condition.}
\vspace{-1.5mm}
\label{tab:corpus-matrix}
\small
\setlength{\tabcolsep}{4.5pt}
\begin{tabular}{@{} l r | ccc | cccccc @{}}
\toprule
& \textbf{Budget} & \multicolumn{3}{c}{\textbf{Formal Sources (M)}} & \multicolumn{6}{c}{\textbf{Informal Sources (M)}} \\
\cmidrule(lr){3-5} \cmidrule(lr){6-11}
\textbf{Mixture Condition} & \textbf{(M)} & BEQ & Wiki & Soleil & CRIFUQ & FB & D\'ep & MR & YT & Reddit \\
\midrule
\emph{Original Corpus} & 86.7 & 15.4 & 31.2 & 5.9 & 0.8 & 2.4 & 22.8 & 4.2 & 2.2 & 1.8 \\
\midrule
Formal-only (100/0) & 30.0 & 8.8 & 17.8 & 3.4 & -- & -- & -- & -- & -- & -- \\
75\% Formal (75/25)   & 30.0 & 6.6 & 13.4 & 2.5 & 0.2 & 0.5 & 5.0 & 0.9 & 0.5 & 0.4 \\
Mixed (50/50)         & 30.0 & 4.4 & 8.9 & 1.7 & 0.4 & 1.1 & 10.0 & 1.8 & 1.0 & 0.8 \\
75\% Informal (25/75) & 30.0 & 2.2 & 4.5 & 0.8 & 0.5 & 1.6 & 15.0 & 2.8 & 1.4 & 1.2 \\
Informal-only (0/100)& 30.0 & -- & -- & -- & 0.7 & 2.1 & 20.0 & 3.7 & 1.9 & 1.6 \\
\bottomrule
\end{tabular}
\end{table*}
\subsection{Register Sweep Across all Models}
\label{sec:register}

\begin{table*}[t]
\centering
\caption{Continuous Register Sweep on CroissantLLM 1.35B: Results are Macro-F1, averaged over three seeds. Perplexity (PPL) is evaluated on held-out Québec French and standard French sets.}
\vspace{-1.5mm}
\label{tab:register}
\footnotesize
\setlength{\tabcolsep}{4pt}
\begin{tabular}{@{} l cc cccc cc @{}}
\toprule
& \multicolumn{2}{c}{\textbf{Québec French}} 
& \multicolumn{4}{c}{\textbf{Standard French}} 
& \multicolumn{2}{c}{\textbf{PPL} $\downarrow$} \\
\cmidrule(lr){2-3} \cmidrule(lr){4-7} \cmidrule(lr){8-9}
\textbf{Condition} & QFrBLiMP & QFrCoRE & AlloCiné & PAWS & Fr-BoolQ & MMS & QC & FR \\
\midrule
\emph{Base (no CPT)} & 89.2 & 27.5 & 50.4 & 35.5 & 56.2 & 25.4 & 17.24 & 8.31 \\
\midrule
Formal-only   & 89.9 & 35.7 & 56.6 & 41.0 & 37.7 & 25.5 & 15.63 & 9.89 \\
75/25         & 89.4 & 33.9 & 54.1 & 39.1 & 40.2 & 25.4 & 15.79 & 9.73 \\
Mixed (50/50) & 90.2 & 33.7 & 50.8 & 39.0 & 40.8 & 25.3 & 15.99 & 9.62 \\
25/75         & 91.0 & 32.2 & 57.2 & 38.6 & 41.7 & 25.3 & 16.30 & 9.53 \\
Informal-only & 88.5 & 28.2 & 60.0 & 40.7 & 38.9 & 25.1 & 23.19 & 10.70 \\
\midrule
Double (60M)$^\dag$ & 90.7 & 33.8 & 61.8 & 39.0 & 43.5 & 25.4 & 15.67 & 9.84 \\
\bottomrule
\multicolumn{9}{@{}l}{\scriptsize $^\dag$60M tokens instead of 30M.}
\end{tabular}
\end{table*}

To evaluate the effect of register, we first examine the formal-only and informal-only variants across all nine LLMs. Full results are in Appendix \ref{app:tables}, in Table~\ref{tab:grammar-lexicon}.

As Figure \ref{fig:extremes} (a) and (b) show, grammatical performance fluctuates minimally, shifting at most 2.0\% on QFrBLiMP and 4.8\% on MultiBLiMP-Fr. This confirms that dialectal CPT safely preserves underlying syntax regardless of register.

Expression comprehension, by contrast, is highly sensitive to corpus composition (see Figure \ref{fig:extremes} (c) and (d)). Formal-only training yields higher QFrCoRE scores across almost all models, averaging a 5.3-point average advantage over informal-only. Conversely, informal training degrades QFrCoRE in eight of nine models, averaging a 4.0\% drop, while raising the standard French equivalent, MFrCoE (5.0\%), in all nine. Formal training drives the exact opposite shift, primarily in weaker base models.

Because QFrCoRE and MFrCoE share identical multiple-choice formats, this opposing shift confirms a change in linguistic knowledge rather than mere format learning. This divergence exposes an evaluation trap: if a model trained on informal Québec data is validated on standard French benchmarks, rising scores create a false illusion of success. Instead, our paired evaluation reveals that target dialect comprehension degrades.

\begin{table*}[t]
\centering
\caption{Leave-One-Genre-Out Ablations at fixed budget for CroissantLLM 1.35B. All rows train on the same 30M token budget. Results are the mean of three seeds.}
\vspace{-2.5mm}
\label{tab:genre}
% \footnotesize
\setlength{\tabcolsep}{3.1pt}
\scalebox{0.89}{
\begin{tabular}{l | cc | cccc | cc}
\toprule
& \multicolumn{2}{c|}{\textbf{Québec French}} & \multicolumn{4}{c|}{\textbf{Standard French}} & \multicolumn{2}{c}{\textbf{PPL $\downarrow$}} \\
\cmidrule(lr){2-3}\cmidrule(lr){4-7}\cmidrule(lr){8-9}
\textbf{Removed Genre} & QFrBLiMP & QFrCoRE & AlloCiné & PAWS & Fr-BoolQ & MMS & QC & FR \\
\midrule
\emph{None (full corpus)} & 90.2 & 33.7 & 53.1 & 39.0 & 40.5 & 25.4 & 15.91 & ~9.68 \\
- Encyclopedic \begin{tiny}{(QC Wiki, 31.2M)}\end{tiny} & 90.1 & 33.8 & 57.3 & 39.7 & 40.3 & 25.4 & 16.32 & ~9.51 \\
- Literary \begin{tiny}{BEQ, (15.4M)}\end{tiny} & 90.3 & 28.3 & 50.8 & 40.3 & 35.2 & 25.3 & 10.46 & 11.73 \\
- News \begin{tiny}{(Le Soleil, 5.9M)}\end{tiny} & 90.2 & 34.8 & 53.5 & 35.9 & 45.8 & 25.3 & 15.85 & ~9.71 \\
- Forums/social \begin{tiny}{(Dépotoir, MtlRacing, FB, Reddit, 31.2M)}\end{tiny} & 89.5 & 34.9 & 65.0 & 38.8 & 41.9 & 25.5 & 15.65 & ~9.85 \\  % leila added scriptsize
- Speech/comments \begin{tiny}{(CRIFUQ, YouTube, 3.0M)}\end{tiny} & 90.0 & 33.7 & 52.9 & 39.2 & 39.7 & 25.4 & 15.89 & ~9.73 \\
\bottomrule
\end{tabular}}
\end{table*}

\subsection{Continuous Register Sweep on CroissantLLM}
\label{sec:croissant-sweep}

To observe the impact of the register across a continuous spectrum, we conduct a register sweep on CroissantLLM 1.35B. We chose this model as our testbed because it shows the largest gain in regional expression comprehension during CPT (see Section \ref{sec:register}). As Table~\ref{tab:register} shows, under formal-only training, its QFrCoRE score rises from a 27.5\% baseline to 35.7\% (+8.2\%). This strong capacity to absorb regional knowledge provides the clear signal needed to accurately track how incremental changes in the data mixture affect dialect acquisition.

When informal text progressively replaces formal text, this dialectal gain decreases monotonically (35.7\% $\to$ 33.9\% $\to$ 7\% $\to$ 32.2\% $\to$ 28.2\%), while grammatical performance (QFrBLiMP) fluctuates in no particular direction. This continuous decline reinforces our broader finding that the target variety's unique vocabulary is sparse in informal data and significantly better preserved in formal sources.

To isolate the effect of register, we verified whether dialectal adaptation relies on specific data curation or simply raw text exposure. If volume were the primary factor, doubling the budget should yield substantial gains. Therefore, we evaluate a 60M-token Mixed Double condition, a 50/50 formal--informal mixture, which reaches a QFrCoRE score of $33.8\%$. As shown in Table~\ref{tab:register}, this is only 0.1\% above the standard 30M-token Mixed (50/50) condition ($33.7\%$). By comparison, allocating the baseline 30M-token budget entirely to formal text increases QFrCoRE to $35.7\%$. Because a targeted 30M-token corpus outperforms the mixed 60M-token corpus, we conclude that data composition matters more than total volume at this scale.

\subsection{Leave-One-Genre-Out Ablations}
\label{sec:composition-ablation}

Having established that formal text drives the acquisition of regional expressions, we use a leave-one-genre-out ablation to verify which genre aids adaptation the most. When a source is removed, the remaining sources are upsampled to maintain the budget, ensuring any performance change is driven by the absence of that genre's linguistic characteristics rather than a reduction in data.

As Table \ref{tab:genre} shows, literary text is the clear driver of dialectal adaptation. Removing 15.4M tokens of BEQ text drops QFrCoRE from the full-corpus reference of 33.7\% to 28.3\%, erasing nearly all CPT gains. Conversely, removing the largest individual source, Québec Wikipedia (31.2M tokens), leaves the score unchanged at 33.8\%.

Grammatical performance (QFrBLiMP), in contrast, is unchanged, fluctuating by less than a single point (89.5\% $\to$ 90.3\%) across all ablations. While the acquisition of new regional expressions is highly sensitive to the genres provided during training, the model's fundamental syntactic capabilities remain preserved regardless of the mixture. QFrBLiMP specifically evaluates the model's robustness against common regional syntactic errors and English calques documented by official Québec linguistic guidelines. Its stability demonstrates that CPT injects divergent dialectal expressions without corrupting the model's pre-existing, shared syntactic foundation.

Beyond dialectal acquisition, the ablation reveals that corpus composition dictates standard French retention. Removing literary text strengthens the decline in standard French reading comprehension (Fr-BoolQ: 40.5\% $\to$ 35.2\%) and degrades standard French perplexity (9.68 $\to$ 11.73). Conversely, removing informal data spikes AlloCiné by 11.9\% (53.1\% to 65.0\%), suggesting that informal text interferes with pre-existing sentiment capabilities.

The perplexity scores further expose a dissociation between language modeling metrics and downstream capability. Removing literary text yields the best Québec French PPL in the entire suite (15.91 $\to$ 10.46), yet this condition produces the worst QFrCoRE score (33.7\% $\to$ 28.3\%). This inverse relationship confirms that perplexity improvements can be driven by simplifying the held-out distribution rather than by acquiring regional features.

Overall, the results in Table \ref{tab:genre} isolate \emph{genre}, not register, to demonstrate that composition matters more than raw volume. Because the budget remains controlled, the decline when removing BEQ (33.7\% $\to$ 28.3\%) confirms that literary text supplies the regional expressions the benchmark requires, whereas removing Wikipedia, a formal encyclopedic source twice the size of BEQ, leaves performance unchanged. This supports the register-sweep result of Section \ref{sec:croissant-sweep}: the benefit of formal-heavy training comes not from formality itself, but from the literary component that formal-heavy mixtures contain more of. We offer a hypothesis: BEQ consists of public-domain Québec literary works, whose older, lexically denser vocabulary makes it the most plausible source of the traditional regional expressions QFrCoRE evaluates, which encyclopedic, news, and modern informal text may rarely contain.

\section{Impact of Base Model Capability}
\label{sec:results-rq2}

Having identified formal-only text as the optimal mixture, we investigate whether dialectal adaptation benefits all LLMs equally. To answer this, we apply the same training procedure across nine LLMs, again using a fixed 30M token adaptation budget. This allows us to verify the role of each LLM's starting competence. Full results are available in Appendix \ref{app:tables}, in Table~\ref{tab:ladder}.

% Figure \ref{fig:ladder} shows a divergence between grammatical acceptability and expression comprehension as model capacity scales. Grammatical performance (QFrBLiMP) remains stable, fluctuating by at most 2.0\% across all nine models regardless of size or initial competence, confirming that syntactic structures are preserved uniformly across the capability spectrum. Regional expression comprehension (QFrCoRE), in contrast, shows a clear inverse relationship between baseline competence and adaptation gains: weaker base models benefit the most. The limited 30M-token budget acts as an equalizer, yielding measurable gains in expression comprehension for smaller models (+8.2\% for CroissantLLM, +1.5\% for Qwen 0.5B), while stronger models hit a capability ceiling, with gains diminishing and even reversing at the top end (-1.4\% for Llama-3.1 8B, -2.2\% for Qwen2.5 14B).

% These results suggest that highly capable models already possess much of the regional knowledge in our corpus, and that a 30M-token signal is too sparse to update their internal representations. Budget-constrained CPT thus effectively bootstraps smaller models, but frontier models likely require larger data volumes or targeted instruction tuning to show measurable dialectal gains.

\begin{figure}[t]
\centering
\begin{tikzpicture}

% --- QFrCoRE (Expressions) ---
\begin{axis}[
  name=core,
  width=\columnwidth,
  height=4.6cm, % Reduced from 5.2cm
  title={\footnotesize (a) Expressions (QFrCoRE)},
  xlabel={\footnotesize Baseline Macro-F1 (unadapted)},
  ylabel={\footnotesize CPT Macro-F1},
  xmin=25, xmax=41,
  ymin=25, ymax=41,
  grid=both,
  grid style={dashed,gray!25},
  tick label style={font=\scriptsize},
  label style={font=\scriptsize},
  legend style={font=\scriptsize, at={(0.05,0.95)}, anchor=north west, draw=gray!50},
]
% Diagonal line (y=x)
\addplot[dashed, thick, gray] coordinates {(25,25) (41,41)};
\addlegendentry{No change ($y=x$)}

% Scatter points
\addplot[
    only marks,
    mark=*,
    blue!70!black,
    nodes near coords,
    point meta=explicit symbolic,
    every node near coord/.append style={font=\tiny, anchor=south west, inner sep=1pt}
] table[meta=label] {
    x y label
    27.5 35.7 {Croissant}
    30.2 31.7 {Qwen 0.5B}
    31.5 32.1 {Qwen 1.5B}
    31.9 33.3 {Llama 1B}
    32.7 34.1 {Qwen 3B}
    33.0 33.1 {Llama 3B}
    33.7 35.2 {Qwen 7B}
    35.9 34.5 {Llama 8B}
    39.1 36.9 {Qwen 14B}
};
\end{axis}

% --- QFrBLiMP (Grammar) ---
\begin{axis}[
  name=blimp,
  at={(core.below south west)}, anchor=above north west, yshift=-0.5cm, % Reduced gap from -0.7cm
  width=\columnwidth,
  height=4.6cm, % Reduced from 5.2cm
  title={\footnotesize (b) Grammar (QFrBLiMP)},
  xlabel={\footnotesize Baseline Macro-F1 (unadapted)},
  ylabel={\footnotesize CPT Macro-F1},
  xmin=82, xmax=92, 
  ymin=82, ymax=92, 
  grid=both,
  grid style={dashed,gray!25},
  tick label style={font=\scriptsize},
  label style={font=\scriptsize},
]
% Diagonal line (y=x)
\addplot[dashed, thick, gray] coordinates {(82,82) (92,92)};

% Scatter points (Calculated as Baseline + Delta)
\addplot[
    only marks,
    mark=square*,
    teal,
    nodes near coords,
    point meta=explicit symbolic,
    every node near coord/.append style={font=\tiny, anchor=south west, inner sep=1pt}
] table[meta=label] {
    x y label
    88.85 89.55 {Croissant}
    84.50 83.80 {Qwen 0.5B}
    86.20 87.60 {Qwen 1.5B}
    83.36 84.16 {Llama 1B}
    90.93 90.53 {Qwen 3B}
    86.77 86.47 {Llama 3B}
    89.04 89.44 {Qwen 7B}
    88.47 87.07 {Llama 8B}
    89.04 89.14 {Qwen 14B}
};
\end{axis}
\end{tikzpicture}
\caption{Impact of Baseline Competence: CPT performance plotted against the unadapted baseline. The diagonal dashed line represents unchanged performance.}
\vspace{-2.5mm}
\label{fig:ladder}
\end{figure}
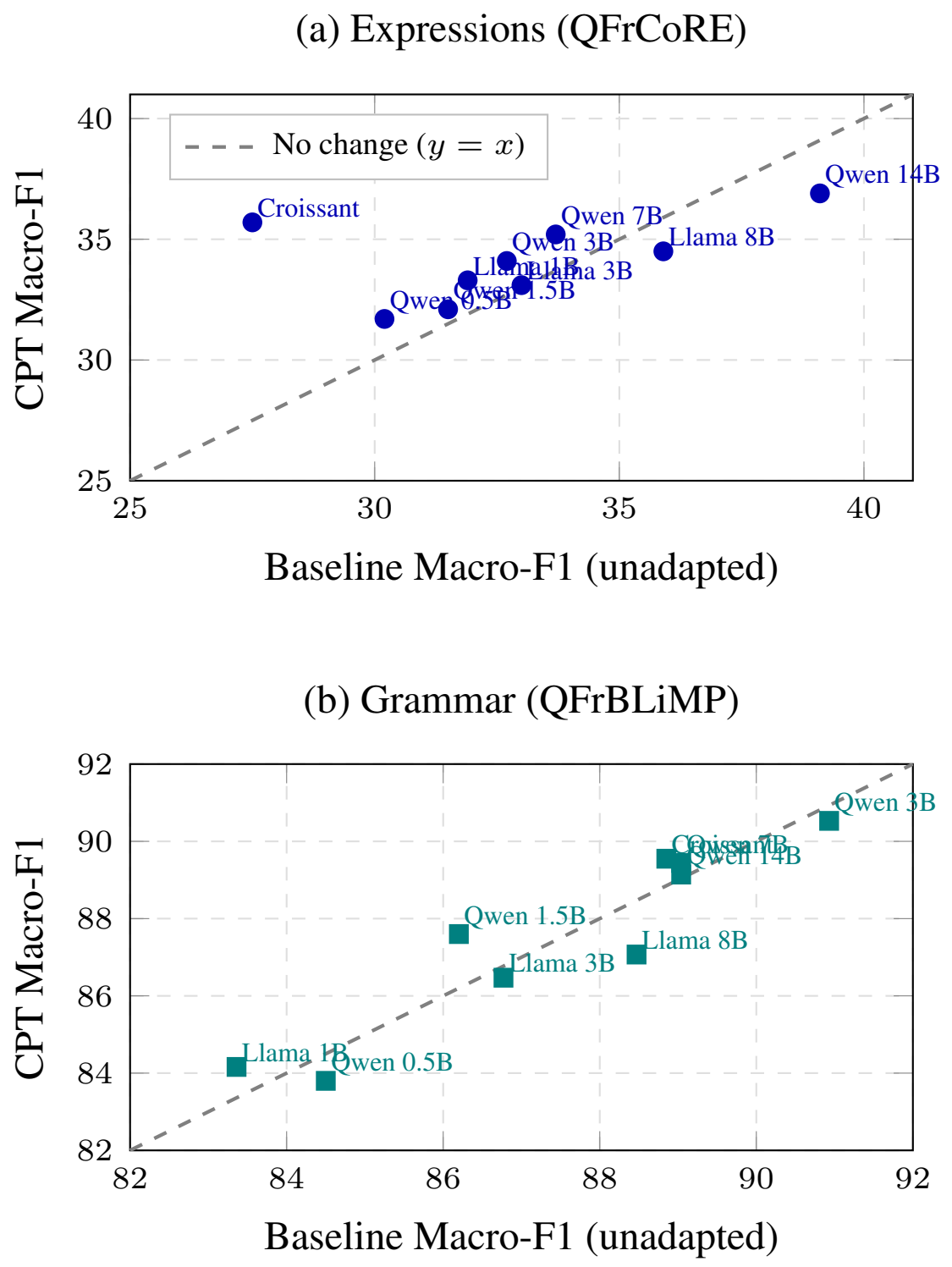

Figure \ref{fig:ladder} shows a divergence between grammatical acceptability and expression comprehension based on a model's initial competence. Plotting CPT performance against the unadapted baseline reveals that grammatical performance (QFrBLiMP) remains stable. Scores fluctuate by at most 2.0\% across all models regardless of size, clustering along the diagonal and confirming that syntactic structures are preserved across the spectrum. 

Regional expression comprehension (QFrCoRE), in contrast, shows an inverse relationship between baseline competence and adaptation gains, where weaker base models benefit the most. The limited 30M-token budget acts as an equalizer, pushing smaller models significantly above the diagonal (+8.2\% for CroissantLLM, +1.5\% for Qwen 0.5B). Meanwhile, improvements in performance shrink as baseline competence increases, and the two strongest baselines fall below the diagonal (−1.4 for Llama-3.1 8B, −2.2 for Qwen2.5 14B). These results suggest that highly capable models already possess much of the regional knowledge in our corpus, and that a 30M-token signal is too sparse to update their internal representations. Budget-constrained CPT thus effectively bootstraps smaller models, but frontier models likely require larger data volumes or targeted instruction tuning to show measurable dialectal gains.

\section{Pitfalls of Aggregate Metrics}
\label{sec:results-rq3}

Finally, to evaluate if aggregate scores conceal forgetting and task-specific trade-offs, we replace a portion of the training budget with standard French replay data from a held-out corpus\footnote{available at \url{https://huggingface.co/datasets/croissantllm/croissant_dataset}.}. This isolates severe, capability-specific regressions and recoveries that would otherwise remain hidden behind stable aggregate averages.

\subsection{Standard French Replay}
\label{sec:costs}
While Section~\ref{sec:results-rq2} showed how base model capability dictates the acquisition of new dialectal knowledge, measuring success solely by these gains ignores the potential capabilities lost in the process. In standard continual learning, adapting a model to a new domain often degrades its prior knowledge, a phenomenon known as catastrophic forgetting \citep{mccloskey1989catastrophic, biesialska-etal-2020-continual}. To expose this hidden cost, we allocate a portion of our 30M-token budget (10\%, 25\%, and 50\%) to standard French replay data. Full results are provided in Appendix \ref{app:tables}, in Table \ref{tab:replay}.

As Figure \ref{fig:replay} shows, evaluating standard French capabilities through a broad aggregate metric makes the replay intervention appear ineffective. Averaging across the four general French benchmarks yields a flat trend line (see the dotted black line), sitting around 3\% below the unadapted baseline (41.9\%) and hovering around 38.8\% regardless of the replay ratio. Relying solely on this macroscopic evaluation would mistakenly lead us to conclude that replay fails to restore any standard French capabilities degraded by dialectal CPT.

Disaggregating these scores, however, reveals opposing capability shifts. As standard French replay increases, performance on reasoning-heavy tasks such as Fr-BoolQ (blue line) recovers monotonically (40.8\% $\to$ 50.1\%). However, as Fr-BoolQ improves by 9.3\%, the standard French sentiment task AlloCiné degrades by 7.7\%. This opposing shift exposes a trade-off under a constrained token budget: re-introducing formal standard French restores logical inference but pulls the model's distribution away from the colloquial language necessary for domain-specific sentiment tasks.

Meanwhile, performance on PAWS-X (green line) and MMS (pink line) remains static across all replay ratios. Because these flat scores are averaged with the opposing swings of Fr-BoolQ and AlloCiné, the resulting aggregate metric is artificially stabilized.

\begin{figure}[t]
\centering
\begin{tikzpicture}
\begin{axis}[
  width=\columnwidth,
  height=5.0cm,
  xlabel={Standard French replay share of the budget (\%)},
  ylabel={PPL $\downarrow$},
  xtick={0,10,25,50},
  ymin=7, ymax=18,
  ymajorgrids,
  grid style={dashed,gray!25},
  tick label style={font=\footnotesize},
  label style={font=\footnotesize},
  legend style={
    font=\scriptsize,
    at={(0.5,0.52)},
    anchor=center,
    legend columns=2,
    draw=gray!50,
    fill=white,
    inner sep=3pt,
    nodes={inner sep=2pt},
    /tikz/every even column/.append style={column sep=0.45cm}
  },
  legend cell align=left,
]

% 1. QC Perplexity
\addplot+[solid, mark=square*, thick, color=orange!80!black, mark options={fill=orange!80!black}] coordinates {(0,15.99) (10,16.02) (25,16.08) (50,16.22)};
\addlegendentry{Qu\'ebec (QC)}

% 2. Standard French Perplexity
\addplot+[solid, mark=*, thick, color=blue!70!black, mark options={fill=blue!70!black}] coordinates {(0,9.62) (10,8.67) (25,8.55) (50,8.47)};
\addlegendentry{Standard (FR)}

\end{axis}
\end{tikzpicture}
\caption{Held-out Perplexity during standard French replay for CroissantLLM 1.35B.}
\vspace{-1mm}
\label{fig:replay-ppl}
\end{figure}
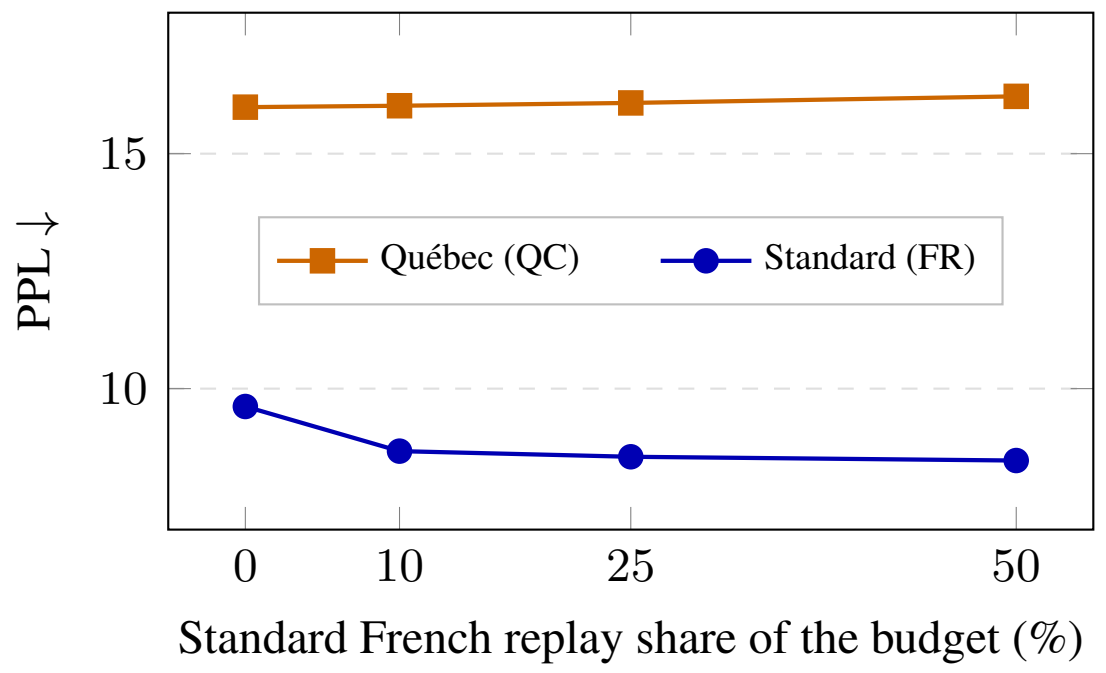

Rather than restoring a generic standard French average, replay specifically rehabilitates reading comprehension without erasing dialectal gains, as QFrCoRE holds steady at 33.7\%. First, as shown in Figure~\ref{fig:replay-ppl}, standard French PPL strictly improves as replay volume increases (9.62 $\to$ 8.47), indicating global language modeling gains despite the stable task average. Second, as reported in Table~\ref{tab:replay} (see Appendix \ref{app:tables}), replacing French-only replay with a balanced French-English source at the 25\% ratio yields nearly identical Fr-BoolQ performance (48.0\% vs. 48.1\%). Because CroissantLLM is bilingual, this suggests that the recovery of its logical reasoning capabilities is not strictly tied to standard French exposure, but can be effectively stimulated by generic pre-training text in either of the model's core languages.

\begin{figure}[t]
\centering
\begin{tikzpicture}
\begin{axis}[
  width=\columnwidth,
  height=5.4cm,
  xlabel={Standard French Replay share of the budget (\%)},
  ylabel={Macro-F1},
  xmin=-20, xmax=55,
  ymin=20, ymax=95,
  xtick={-15, 0, 10, 25, 50},
  xticklabels={Base, 0, 10, 25, 50},
  ymajorgrids,
  grid style={dashed,gray!25},
  tick label style={font=\footnotesize},
  label style={font=\footnotesize},
legend style={
  font=\tiny,
  at={(0.5,0.70)},
  anchor=center,
  legend columns=2,
  draw=gray!50,
  fill=white,
  inner sep=2pt,
  nodes={inner sep=1pt},
  /tikz/every even column/.append style={column sep=0.25cm}
},
  legend cell align=left,
]

% Séparateur visuel entre la "Base" et les expériences de Replay
\draw[densely dotted, thick, gray] (-7.5, 20) -- (-7.5, 95);

% 1. AlloCiné (Base: 50.4)
\addplot+[mark=triangle*, thick, color=red!60!black, mark options={fill=red!60!black}] coordinates {(-15,50.4) (0,50.8) (10,47.7) (25,45.4) (50,43.1)};
\addlegendentry{AlloCiné}

% 2. Fr-BoolQ (Base: 56.2)
\addplot+[mark=*, thick, color=blue!70!black, mark options={fill=blue!70!black}] coordinates {(-15,56.2) (0,40.8) (10,45.8) (25,48.1) (50,50.1)};
\addlegendentry{Fr-BoolQ}

% 3. Aggregate Standard French Average (Base: ~41.9)
\addplot+[mark=diamond*, thick, densely dashed, color=darkgray, mark options={fill=darkgray}] coordinates {(-15,41.9) (0,39.0) (10,38.8) (25,38.7) (50,38.6)};
\addlegendentry{Std-Fr Average}

% 4. QFrCoRE (Base: 27.5)
\addplot+[mark=square*, thick, color=orange!80!black, mark options={fill=orange!80!black}] coordinates {(-15,27.5) (0,33.3) (10,33.9) (25,33.7) (50,33.7)};
\addlegendentry{QFrCoRE (QC)}

% 5. PAWS-X (Base: 35.5)
\addplot+[solid, mark=pentagon*, thick, color=green!50!black, mark options={fill=green!50!black}] coordinates {(-15,35.5) (0, 39.0) (10, 36.2) (25, 36.0) (50, 35.7)};
\addlegendentry{PAWS-X}

% 6. MMS (Base: 25.4)
\addplot+[solid, mark=otimes*, thick, color=magenta!70!black, mark options={fill=magenta!70!black}] coordinates {(-15,25.4) (0, 25.3) (10, 25.4) (25, 25.4) (50, 25.4)};
\addlegendentry{MMS}

% 7. QFrBLiMP (Base: 89.2)
\addplot+[solid, mark=star, thick, color=teal, mark options={fill=teal}] coordinates {(-15,89.2) (0,90.2) (10,90.0) (25,90.3) (50,89.8)};
\addlegendentry{QFrBLiMP (QC)}

\end{axis}
\end{tikzpicture}
\caption{Standard French replay on the 50/50 Québec French mixture for CroissantLLM 1.35B.}
\vspace{-1mm}
\label{fig:replay}
\end{figure}
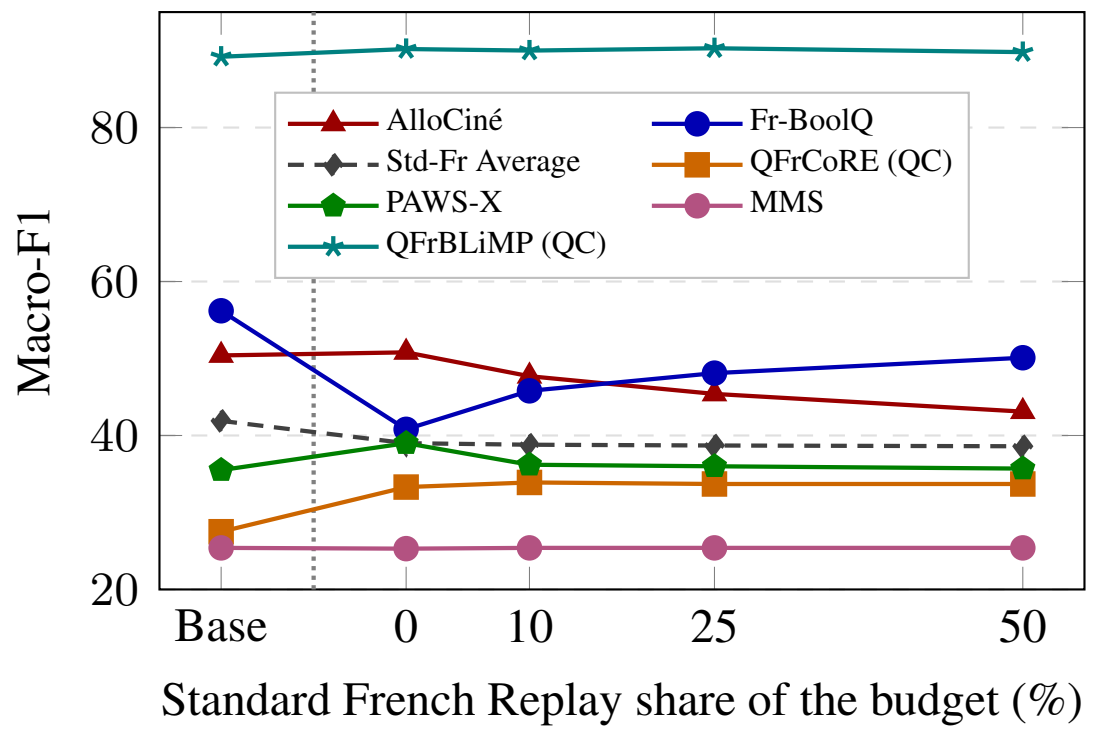

\subsection{Aggregate Metrics Mislead}
\label{sec:measurement}

The masking effect observed in Section~\ref{sec:costs} exposes an important flaw in aggregate scoring: broad averages conceal severe capability trade-offs forced by replay under a constrained token budget. This masking also occurs during adaptation itself. As noted in Section~\ref{sec:register}, the formal-only condition improves overall Québec dialect performance while hiding an 18.5\% degradation on Fr-BoolQ (56.2\% baseline $\to$ 37.7\%).

Beyond aggregate averages, held-out perplexity presents a second evaluation trap. When dedicated dialectal benchmarks are unavailable, perplexity is often used as a proxy for adaptation success \citep{gururangan-etal-2020-dont, wettig2024qurating}. Our results caution against this substitution. The clearest evidence is the within-model dissociation in the genre ablation (see Table~\ref{tab:genre}): the experiment with literary text removed achieves the best (lowest) PPL$_{\mathrm{QC}}$ of any CroissantLLM condition (15.91 $\to$ 10.46) yet yields the worst QFrCoRE score (28.3\%).

Removing lexically dense literary text from the training corpus simplifies the language modeling task, which artificially improves perplexity. However, because this literary text is the primary source of regional vocabulary, removing it erases the model's ability to comprehend the dialectal expressions measured by the benchmark. While perplexity is not always inversely correlated with downstream performance, this dissociation proves it is an unreliable predictor of dialectal acquisition: a model can achieve "better" perplexity simply by training on standard, predictable text, even as it fails to learn the target dialect.

Echoing recent critiques of NLP leaderboards \citep{ethayarajh-jurafsky-2020-utility, bowman-dahl-2021-will}, our findings demonstrate that shortcut metrics, whether aggregate averages or perplexity scores, are insufficient for evaluating dialectal adaptation in a comprehensive way. Validating adaptation requires the disaggregated, per-task reporting adopted throughout this paper.

\section{Conclusion}
\label{sec:conclusion}

We presented a controlled study of dialectal CPT for Qu\'ebec French language understanding on nine LLMs from three families. By restricting the training budget to 30M tokens, we demonstrated that the efficacy of adaptation is dictated primarily by a model's baseline competence rather than the process itself. While our weakest base model gained 8.2 points in regional expression comprehension, stronger models hit a capability ceiling, yielding no benefit, leaving the most capable unadapted models strictly superior to all adapted variants.

Furthermore, when CPT succeeds, its benefits are narrow: improvements are concentrated in expression comprehension rather than grammar, driven by formal literary text rather than raw volume, and highly sensitive to register. Notably, relying on informal genres actively degraded target dialect comprehension in eight of the nine models. Finally, we exposed the zero-sum nature of these gains in expression comprehension: severe degradations in reading comprehension, which standard French replay recovers but aggregate task averages conceal.

Additionally, our results caution against using held-out perplexity as a proxy for adaptation success. We reveal a dissociation where simplifying the held-out distribution, such as removing lexically dense literary text, artificially improves perplexity while destroying the model's acquisition of regional expressions.

Ultimately, our findings challenge the assumption that dialectal CPT is a uniformly beneficial default. Effective low-resource adaptation requires assessing the base model's competence, curating adaptation corpora by genre rather than volume, and favoring disaggregated, task-level evaluation.

\section*{Limitations}

Although this study highlights important gaps in the assessment of dialectal adaptation, our empirical scope is limited. Our analysis is restricted to Québec French and relies exclusively on the COLE benchmark suite alongside its standard French counterparts. Consequently, while our methodological critiques regarding aggregate metrics and perplexity are applicable to the Québec French dialect, the specific dynamics we observe, such as the strong impact of literary text, must be validated across other dialects and languages before being generalized.

Secondly, all models are adapted with the same LoRA configuration (see Appendix~\ref{sec:hyperparameters}). We fixed these values so that differences in performance could be attributed to the base model and the corpus rather than to tuning, and because of hardware constraints. This configuration produces relatively small updates and may underestimate what CPT can achieve for larger and stronger models, so the absence of gains for these models should be read as a property of this setting rather than a theoretical upper bound. We do not evaluate full-parameter training, other LoRA ranks or learning rates, or budgets beyond 60M tokens.

Additionally, our evaluation covers language understanding only. Dialectal adaptation is often motivated by the wish for models that produce the target variety, and our results do not show whether CPT changes the dialectal character of generated text. Evaluating generation, for instance through human judgments of Québec French expressions and constructions in model outputs, is an important direction for future work.

% Additionally, our model training is done using LoRA at a single rank under a strict 30M token budget. While this budget realistically mirrors the data scarcity inherent to low-resource dialects, we do not evaluate full-parameter fine-tuning or scaling laws beyond this threshold. 

Finally, while we show that formal literary text drives the acquisition of regional expressions, the exact reason why this occurs has not been investigated in this study. Future work should analyze these literary texts more deeply to understand why they so heavily outperform web data for dialectal adaptation.

% \section{Acknowledgements}
% put in the final version, not submitted version for double peer review

\section*{Acknowledgments}
This project was undertaken thanks to funding from NSERC, IVADO and the Canada First Research Excellence Fund. 
The authors would also like to thank CN2i / Le Soleil for generously giving us access to their data.

\bibliography{custom}

\appendix

\clearpage
\section{Training Details and Hyperparameters}
\label{sec:hyperparameters}

All experiments are implemented using LoRA with rank $r=32$, scaling factor $\alpha=64$, and dropout $0$. LoRA is applied to all attention and feed-forward projections ($q$, $k$, $v$, $o$, $\mathrm{gate}$, $\mathrm{up}$, and $\mathrm{down}$), while the base-model weights are completely frozen.

We train using the AdamW optimizer with a weight decay of $0.01$, a peak learning rate of $1\mathrm{e}{-5}$, and a cosine learning rate schedule with a $0.1$ warmup ratio. We apply gradient clipping at $1.0$, use a sequence length of 1024 with a stride of 512, and train using mixed precision (fp16) with gradient checkpointing. We reserve 5\% of each source for validation and held-out perplexity evaluation; this text is never used for training. At the end of training, we select the final model checkpoint based on the lowest validation loss.

\vfill\eject

\section{Full Result Tables}
\label{app:tables}

\subsection{Register Sweep Across all Models}
\begin{table}[h]
\centering
\caption{Grammatical performance and expression comprehension under the two register extremes (formal-only and informal-only CPT), for all nine models. Results are reported as the change in Macro-F1 relative to each model's unadapted baseline.}
\label{tab:grammar-lexicon}
%\small
\setlength{\tabcolsep}{4.5pt}
\resizebox{\columnwidth}{!}{%
\begin{tabular}{@{} l rrrr rrrr @{}}
\toprule
& \multicolumn{4}{c}{\textbf{Grammar}} & \multicolumn{4}{c}{\textbf{Expressions}} \\
\cmidrule(lr){2-5} \cmidrule(lr){6-9}
& \multicolumn{2}{c}{\textbf{QFrBLiMP}} & \multicolumn{2}{c}{\textbf{MultiBLiMP}-Fr} & \multicolumn{2}{c}{\textbf{QFrCoRE}} & \multicolumn{2}{c}{\textbf{MFrCoE}} \\
\cmidrule(lr){2-3} \cmidrule(lr){4-5} \cmidrule(lr){6-7} \cmidrule(lr){8-9}
\textbf{Model} & Formal & Inf. & Formal & Inf. & Formal & Inf. & Formal & Inf. \\
\midrule
Croissant 1.35B & $+0.7$ & $-0.8$ & $-1.3$ & $-1.3$ & $+8.2$ & $+0.7$ & $+2.0$ & $+4.8$ \\
Qwen2.5 0.5B    & $-0.7$ & $-2.0$ & $-4.8$ & $+1.7$ & $+1.5$ & $-3.4$ & $+0.7$ & $+4.2$ \\
Qwen2.5 1.5B    & $+1.4$ & $+0.4$ & $+1.7$ & $+1.3$ & $+0.6$ & $-4.0$ & $+0.3$ & $+5.3$ \\
Llama-3.2 1B    & $+0.8$ & $-0.4$ & $-0.9$ & $-2.2$ & $+1.4$ & $-4.3$ & $+0.4$ & $+2.2$ \\
Qwen2.5 3B      & $-0.4$ & $-1.3$ & $0.0$ & $+0.9$ & $+1.4$ & $-4.3$ & $-1.1$ & $+6.6$ \\
Llama-3.2 3B    & $-0.3$ & $-0.7$ & $-1.3$ & $-2.2$ & $+0.1$ & $-6.1$ & $+0.1$ & $+3.8$ \\
Llama-3.1 8B    & $-1.4$ & $+0.1$ & $0.0$ & $-0.1$ & $-1.4$ & $-3.3$ & $-0.3$ & $+5.7$ \\
Qwen2.5 7B      & $+0.4$ & $0.0$ & $0.0$ & $0.0$ & $+1.5$ & $-3.8$ & $-0.2$ & $+6.1$ \\
Qwen2.5 14B     & $-0.7$ & $+0.1$ & $-1.7$ & $-3.2$ & $-2.2$ & $-7.8$ & $-0.9$ & $+6.5$ \\
\bottomrule
\end{tabular}%
}
\end{table}

\subsection{Impact of Base Model Capability}

\begin{table}[h]
\centering
\caption{Formal-only CPT versus each model's own untrained baseline on QFrCoRE. Scores are reported as Macro-F1, and taken as the mean of three random seeds.}
\label{tab:ladder}
%\small
\setlength{\tabcolsep}{4pt}
\scalebox{.87}{
\begin{tabular}{l l ccc}
\toprule
\textbf{Model} & \textbf{Family} & \textbf{Base} & \textbf{$+$CPT} & $\Delta$ \\
\midrule
CroissantLLM 1.35B & Croissant & 27.5 & 35.7 & $+8.2$ \\
Qwen2.5 0.5B & Qwen & 30.2 & 31.7 & $+1.5$ \\
Qwen2.5 1.5B & Qwen & 31.5 & 32.1 & $+0.6$ \\
Llama-3.2 1B & Llama & 31.9 & 33.3 & $+1.4$ \\
Llama-3.2 3B & Llama & 33.0 & 33.1 & $+0.1$ \\
Llama-3.1 8B & Llama & 35.9 & 34.5 & $-1.4$ \\
Qwen2.5 3B & Qwen & 32.7 & 34.1 & $+1.4$ \\
Qwen2.5 7B & Qwen & 33.7 & 35.2 & $+1.5$ \\
Qwen2.5 14B & Qwen & 39.1 & 36.9 & $-2.2$ \\
\bottomrule
\end{tabular}
}
\end{table}

\subsection{Standard French Replay}

\begin{table}[h]
\centering
\caption{Standard French replay on a 50/50 Québec French mixture for CroissantLLM 1.35B. Deltas discussed in the text are relative to the unadapted base model in Table~\ref{tab:register}. Results are the mean of three seeds.}
\label{tab:replay}
\resizebox{\columnwidth}{!}{%
\begin{tabular}{@{} l cc cccc cc @{}}
\toprule
& \multicolumn{2}{c}{\textbf{Québec French}} 
& \multicolumn{4}{c}{\textbf{Standard French}} 
& \multicolumn{2}{c}{\textbf{PPL} $\downarrow$} \\
\cmidrule(lr){2-3}\cmidrule(lr){4-7}\cmidrule(lr){8-9}
\textbf{Condition} & \textbf{BLiMP} & \textbf{CoRE} & \textbf{AlloCiné} & \textbf{PAWS} & \textbf{BoolQ} & \textbf{MMS} & \textbf{QC} & \textbf{FR} \\
\midrule
No replay (ref.) & 90.2 & 33.3 & 50.8 & 39.0 & 40.8 & 25.3 & 15.99 & 9.62 \\
10\% replay      & 90.0 & 33.9 & 47.7 & 36.2 & 45.8 & 25.4 & 16.02 & 8.67 \\
25\% replay      & 90.3 & 33.7 & 45.4 & 36.0 & 48.1 & 25.4 & 16.08 & 8.55 \\
50\% replay      & 89.8 & 33.7 & 43.1 & 35.7 & 50.1 & 25.4 & 16.22 & 8.47 \\
25\% FR+EN       & 90.4 & 33.7 & 45.7 & 36.0 & 48.0 & 25.4 & 16.08 & 8.55 \\
\bottomrule
\end{tabular}%
}
\end{table}

\end{document}